\documentclass{article}

\usepackage{PRIMEarxiv}

\usepackage[utf8]{inputenc} 
\usepackage[T1]{fontenc}    
\usepackage{hyperref}       
\usepackage{url}            
\usepackage{booktabs}       
\usepackage{amsfonts}       
\usepackage{fancyhdr}       
\usepackage{graphicx}       
\graphicspath{{./}{figures/}{figures_v2/}{media/}} 
\usepackage{siunitx}
\usepackage{xcolor}
\usepackage{subcaption}
\usepackage{bm}
\usepackage{mhchem}
\usepackage{amsmath}
\usepackage{makecell}
\usepackage{float}          
\usepackage{placeins}       

\title{A Reinforcement-Learning-Augmented Liquid-Fueled Reactor Network Model for Predicting Lean Blowout in Gas Turbine Combustors}

\author{
  Philip John\textsuperscript{1}, Eloghosa Ikponmwoba\textsuperscript{1}, Pinaki Pal\textsuperscript{2}, Opeoluwa Owoyele\textsuperscript{1}%
  \thanks{Corresponding author: \texttt{oowoyele@lsu.edu}} \\[3pt]
  \textsuperscript{1}Department of Mechanical and Industrial Engineering \\
  Louisiana State University, Baton Rouge, LA 70803, USA \\[3pt]
  \textsuperscript{2}Transportation and Power Systems Division \\
  Argonne National Laboratory, 9700 S. Cass Ave, Lemont, IL 60439, USA
}

\begin{document}
\maketitle

\begin{abstract}
This study introduces a reinforcement learning (RL) framework for generating optimal liquid-fueled reactors to improve lean blowout (LBO) predictions in gas turbine combustors. Existing approaches for determining cluster boundaries rely on manual heuristics or distance-based metrics in the input space. In contrast, the proposed method is goal-oriented, explicitly accounting for the target metric (e.g., LBO prediction accuracy) during cluster formation. The framework employs a multi-stage clustering--classification strategy: an initial clustering step (e.g., $k$-means clustering) generates a large set of homogeneous micro-clusters, followed by an actor-critic RL agent that merges them into optimal reactor zones. The validation study, performed using a Jet-A mechanism (119 species, 841 reactions), shows the RL framework offers improved predictive fidelity compared to $k$-means and captures the correct LBO trends, while achieving substantial speedups relative to the high-fidelity computational model. Overall, the RL-driven approach demonstrates strong potential as a computationally efficient reduced-order modeling technique that can complement high-fidelity simulations for rapid design-space exploration.
\end{abstract}


\section*{Novelty and Significance Statement}
This study presents the first framework that integrates reinforcement-learning decision policies into chemical reactor network construction, enabling reactor partitions and volumes to adapt based on physics-level objectives rather than distance-based criteria. This also represents the first demonstration of a liquid-fueled reactor network capable of reproducing parametric LBO trends across varying operating conditions. The significance of this work lies in the ability of the proposed framework to produce physically consistent LBO trends at a fraction of the computational cost of CFD, a capability that existing reduced-order methods fail to provide. This makes the framework a practical complement to high-fidelity CFD for rapid design-space exploration of gas turbine combustors operating near the LBO limit.

\section{Introduction}
\label{sec:introduction}
Lean blowout (LBO) represents a critical operational limit in gas turbine combustors, particularly for systems operating at lean equivalence ratios \cite{gupta2019prevention, kaluri2018realtime, cheng2024large}. In liquid-fueled systems, the underlying physics are additionally influenced by droplet evaporation and fuel-air mixing processes, which introduce additional timescales prior to chemical reaction \cite{FAETH19831}. Accurate characterization of the LBO limit has traditionally relied on experimental diagnostics such as chemiluminescence imaging and direct flame visualization \cite{STOHR20112953}. More recently, high-fidelity computational fluid dynamics (CFD) approaches have emerged as powerful tools for studying LBO \cite{esclapez2017large, endo2021numerical, dasgupta2024fuel, nassini2021lean}, enabling detailed resolution of the flow, spray, and combustion fields, as well as virtual testing of fuel and geometric effects. However, such approaches remain computationally expensive, particularly for large eddy simulations (LES) of reacting two-phase flows \cite{BOILEAU20082,JARAVEL2018180}, limiting their practicality for large parametric studies and combustor design optimization.

In response to these costs, chemical reactor networks (CRNs) have emerged as computationally efficient alternatives for modeling gas turbines and other combustion systems \cite{falcitelli2002modelling, park2013prediction}. CRN modeling involves partitioning the CFD domain into individual reactor zones, followed by the application of simplified governing equations for each reactor \cite{hataysal2016coupled, falcitelli2002modelling}. Several approaches have been developed to obtain these partitions. Early approaches relied on manual partitioning, wherein an expert delineates the flow-fields into zones based on \textit{a priori} knowledge of the flow and combustion physics \cite{andreini2004gas, novosselov2006chemical, kaluri2018realtime}. To reduce dependence on expert intervention, automated approaches have been introduced, including threshold-based filtering \cite{fichet2010reactor, Grimm2022, perpignan2019modeling}, and more recently, clustering-based methods that group computational cells by thermochemical similarity \cite{savarese2023machine}. However, the application of CRNs to liquid-fueled systems introduces additional modeling challenges that are largely absent in gas-fueled configurations \cite{john2025liquid}. Specifically, the presence of a dispersed liquid phase requires the reactor topology to accommodate not only gas-phase thermochemical variation, but also spray-phase phenomena including droplet breakup, evaporation, and the spatial evolution of fuel-air mixing.
\begin{figure*}[t]
    \centering
    \includegraphics[width=\textwidth]{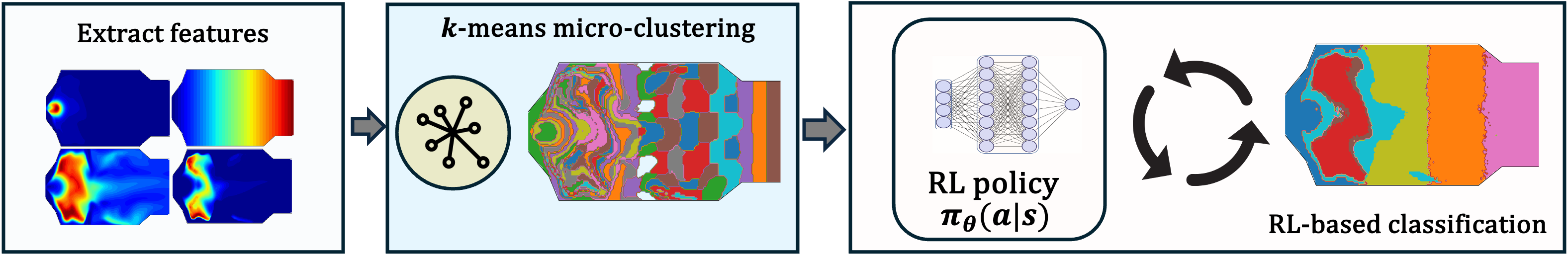}
    \caption{Schematic of the RL--augmented liquid-fueled reactor network framework.}
    \label{fig:abstract_graphics}
\end{figure*}
A common limitation of existing automated CRN clustering approaches is that the reactor partitions are constructed solely from distances in the input feature space, without considering their impact on the target quantity of interest. Because quantities such as \ce{NOx} and LBO depend nonlinearly on the underlying thermochemical variables, distance-based clustering can produce suboptimal reactor configurations. This limitation is particularly severe for LBO, where small changes in the primary reaction zone can strongly affect extinction behavior. Consequently, prior CRN-based LBO studies have largely been validated only at isolated operating conditions \cite{xiao2019predicting, kaluri2018realtime}, with limited demonstration of trend prediction across parametric variations.

In this study, we address this limitation by introducing a reinforcement-learning (RL) framework for automated extraction of optimal reactors in a liquid-fueled reactor network (LFRN) aimed at minimizing LBO prediction error. Unlike distance-based clustering approaches, the RL agent explores reactor partitions and is rewarded directly based on LBO predictive accuracy, thereby promoting the discovery of physically relevant partitions that conventional feature-space metrics may fail to capture. The approach is illustrated in Fig.~\ref{fig:abstract_graphics}. The main contributions are: (1) development of an RL-based framework for goal-oriented reactor-network generation; (2) validation for liquid-fueled LBO prediction using a 119-species Jet-A mechanism \cite{XU2018520,WANG2018502}; and (3) quantitative demonstration of improved performance relative to conventional distance-based clustering. The trained reactor network is intended to be locally predictive within the combustor geometry and operating envelope represented by the baseline CFD data, consistent with standard CRN models discussed above. However, the framework itself is general and can be reapplied to other configurations following the same procedure.

\section{Methodology}
\label{sec:methodology}

\subsection{CFD Numerical Setup and Computational Domain}
\label{sec:cfd_methodology}
\begin{figure}[H]
\centering
\includegraphics[width=0.4\textwidth]{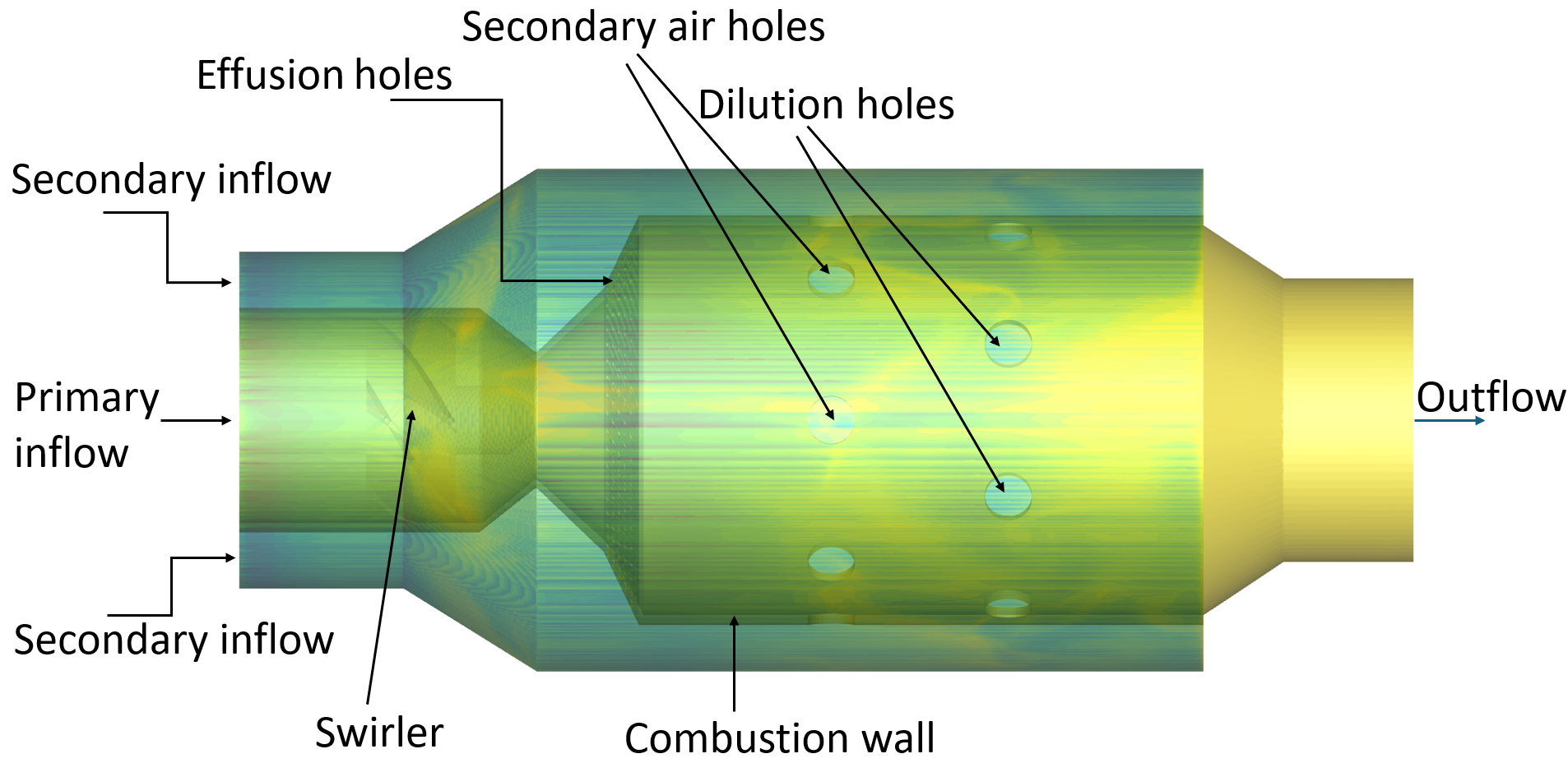}
\caption{Liquid-fueled swirl-stabilized combustor geometry.}
\label{fig:geometry}
\end{figure}

The combustor geometry (Fig.~\ref{fig:geometry}) features a cylindrical can combustor (radius 0.03~m, length 0.2~m) with a single outlet, primary and secondary air inlets, and secondary/dilution holes (8~mm diameter) distributed along the liner walls. A central fuel injector (250~$\mu$m diameter) is surrounded by a six-vane swirl-stabilized primary inlet that generates a central recirculation zone (CRZ) for flame stabilization. Air enters at 350~K through the primary and secondary inlets, with 25\% of the secondary inflow going through the effusion holes and the remaining 75\% going through the secondary and dilution holes. The resulting primary equivalence ratio is $0.35$. A Rosin--Rammler distribution with $d_{32}=20~\mu$m \cite{de2021soot,ECKEL2019134} and spread parameter $n=2.8$ was employed for the spray injection, with a cone half-angle of $35^\circ$ and injection velocity of 63~m/s. Primary/secondary air and dilution holes were modeled as mass-flow inlets, the outlet as a pressure outlet, and the walls as no-slip adiabatic boundaries. Additional details are presented in Table~\ref{tab:bc}.

\begin{table}[H]
\centering
\caption{Baseline operating and boundary conditions.}
\label{tab:bc}
\footnotesize
\setlength{\tabcolsep}{6pt}
\renewcommand{\arraystretch}{1.15}
\begin{tabular}{@{}lr@{}}
\hline
Parameter & Value \\
\hline
Combustor radius & 0.03 m \\
Combustor length & 0.2 m \\
Fuel injector diameter & 250 $\mu$m \\
Operating pressure & 2 atm \\
Air inlet temperature & 350 K \\
Fuel inlet temperature & 327 K \\
Primary air flow rate & 0.0487 kg/s \\
Secondary air flow rate & 0.1135 kg/s \\
Fuel mass flow rate & 0.0012 kg/s \\
Global equiv.\ ratio ($\phi_{\text{global}}$) & 0.35 \\
\hline
\end{tabular}
\end{table}

The Flamelet Generated Manifold (FGM) approach \cite{van2000flamelet} was employed to model turbulence-chemistry interactions, pre-computing chemistry using flamelet equations in mixture fraction space \cite{peters2000turbulent} to generate a lookup table parameterized by a filtered progress variable $\widetilde{c}$, defined as the normalized sum $c = (Y_{\text{CO}_2} + Y_{\text{CO}})/(Y_{\text{CO}_2}^{\text{eq}} + Y_{\text{CO}}^{\text{eq}})$, where superscript ``eq'' denotes equilibrium values, mixture fraction $\widetilde{Z}$, mixture fraction variance $\widetilde{Z''^2}$, and enthalpy $\widetilde{h}$. The flamelet library was generated for Jet-A using the HyChem kinetic mechanism (119 species, 841 reactions) \cite{XU2018520,WANG2018502}. Liquid spray was modeled using Lagrangian particle tracking with the Taylor Analogy Breakup (TAB) model \cite{orourke1987} for breakup, Frossling correlation \cite{amsden1989} for evaporation, and Rosin-Rammler distribution \cite{rosinrammler1933} for initial droplet size.

A Reynolds-Averaged Navier-Stokes (RANS) $k$--$\varepsilon$ simulation was first conducted at a baseline condition to calibrate the reactor network (RN) model as done in other studies \cite{falcitelli2002modelling,savarese2023machine,park2013prediction,hataysal2016coupled,novosselov2006chemical}, thus providing the mean flow and thermochemical fields used for reactor zone construction. Parametric Large Eddy Simulations (LES) (using a dynamic structure model \cite{pomraning2000development}) were subsequently performed across a range of operating conditions to predict lean blowout (LBO) equivalence ratios. A baseline LES is used for training the RL model, while other conditions are used for comparison with the CRNs. The CFD simulations were fully 3-dimensional and performed using CONVERGE CFD (v5.0.2) \cite{richards2025converge}. The RANS and LES meshes contained approximately 2 million and 3 million cells, respectively, with the LES count corresponding to the peak adaptive mesh refinement (AMR) level. The LES employed AMR to refine regions of high velocity and temperature gradients, with the grid dynamically refined from a 2.5~mm base size to a minimum cell size of 0.625~mm in the primary reaction zone. The RANS used a reduced base grid of 2.0~mm which was refined to a minimum of 1.0~mm in the reaction zone. AMR was activated after an initial flow-through time to allow the bulk flow to develop before refinement. A grid convergence study showed that further refinements (to the base grid or AMR level) did not alter the predicted blowout equivalence ratio. The LES simulations utilized 256 processors and required approximately 2--4 days per operating condition.

\subsection{Reactor Network Development}
\label{sec:dev_rn}
Unlike CFD simulations that discretize the domain into millions of cells, the reactor network methodology partitions the combustor volume into a relatively small number of zones, with each described by simplified transport equations. In this study, we employ standard reactor types, including perfectly stirred reactors (PSRs) and plug flow reactors (PFRs), alongside two specialized reactors introduced to account for the liquid-fuel physics. The reactor network is solved using Cantera (v3.0)\cite{cantera2022} with its in-built reactor classes. The plug flow zone is solved along the axial coordinate (rather than as a chain of PSRs), and all reactors are solved at steady state. The choice of reactors is determined by considering the dominant physical processes in each reactor.

A PSR assumes perfect spatial mixing within the reactor volume $V$, so that all properties are uniform in space and may only vary in time. A PFR, in contrast, permits spatial gradients only along the axial flow direction $z$. The equations used to model the PSR and PFR are based on the standard equations used in combustion literature, and can be found in a previous study by the authors \cite{john2025liquid}. Inter-reactor mass flow rates are determined from the RANS simulation of the combustor. To capture the two-phase physics specific to liquid-fueled combustors, two additional specialized reactors are introduced: an \textit{evaporator/breakup reactor} and a \textit{mixing reactor}. Further details can be found in a previous study by the authors \cite{john2025liquid}. In the evaporator/breakup reactor, the gas-phase species and energy equations are augmented with source terms that account for the transfer of mass and energy from the droplet phase to the gas phase. The species conservation equation becomes:
\begin{equation}
\dot{m}_{\text{in}}(Y_{k,\text{in}} - Y_k)
+ \dot{\omega}_k W_k V
+ \dot{S}_k^{\text{evap}} V = 0
\label{eq:evap_species}
\end{equation}
where $\dot{S}_k^{\text{evap}}$ is the evaporative mass source of fuel vapor from the droplet phase, computed from the Frossling evaporation correlation consistent with the CFD setup. The corresponding energy equation includes a spray enthalpy source term to account for the latent heat consumed during evaporation and the sensible heat carried by the evaporating fuel:
\begin{equation}
\begin{gathered}
\dot{m}_{\text{in}} \sum_{k} Y_{k,\text{in}}
h_k(T_{\text{in}}) - \dot{m}_{\text{out}}
\sum_{k} Y_k h_k(T) \\
+ \dot{S}_e^{\text{evap}} V = 0,
\end{gathered}
\label{eq:evap_energy}
\end{equation}
where $\dot{S}_e^{\text{evap}}$ is the net energy source from the evaporating droplets, which includes the latent heat of vaporization and the sensible enthalpy of the released fuel vapor. The evaporator/breakup reactor also tracks the evolution of the droplet size distribution through the breakup process, consistent with the TAB breakup model used in the CFD, providing an updated mean droplet diameter as input to the downstream mixing reactor. The mixing reactor receives the partially evaporated spray and models the continued evaporation of the remaining liquid fuel and the subsequent gas-phase fuel-air mixing prior to combustion. It employs the same augmented species and energy equations (Eqs.~\ref{eq:evap_species}--\ref{eq:evap_energy}), with the evaporation source terms now representing the residual liquid fuel that did not evaporate in the upstream stage. Thus, the evaporation/breakup reactor captures breakup and evaporation, while the mixer captures only further evaporation and subsequent mixing. The outlet of the mixing reactor provides the fully gas-phase inlet conditions, including fuel vapor concentration, mixture fraction, and temperature, for the downstream PSR and PFR combustion zones. This sequential arrangement ensures that the overall reactor topology captures the physical order of events, namely atomization, evaporation, mixing, and combustion, that governs LBO behavior in liquid-fueled combustion systems. All the reactor networks used the same mechanism as the CFD simulations.

\subsubsection{$k$-means Clustering}
\label{sec:clustering_kmeans}
Reactor network construction begins by extracting flow and thermochemical information from three-dimensional CFD results at representative operating conditions. Each computational cell is characterized by a feature vector comprising temperature, equivalence ratio, \ce{OH} mass fraction, and streamwise distance $x$: $\bm{\theta}_i = [T_i,\, \phi_i,\, Y_{\text{OH},i},\, x_i].$ The full dataset is organized in matrix form, $\bm{\Theta} = [\bm{\theta}_1^T, \bm{\theta}_2^T, \ldots, \bm{\theta}_{n_o}^T]^T$, with $n_o$ denoting the total number of CFD cells. Specifically, $T$ and $Y_{\mathrm{OH}}$ characterize local reactivity and heat-release activity, $\phi$ captures the local mixing state, and the streamwise coordinate $x$ enforces spatial coherence of the reactor partition. Since these variables were sufficient to capture the observed LBO behavior, the feature set was not expanded further.

To maintain computational tractability with the large CFD dataset, which contains on the order of $10^6$ points, a hierarchical clustering strategy is used. First, the domain is partitioned into $n_f = 70$ \emph{micro-clusters} using $k$-means, with each micro-cluster represented by its centroid in feature space. Using the RL approach described in Section \ref{sec:clustering_rl}, these 70 centroids are then mapped to a smaller set of $n_c = 7$ larger clusters that define the final reactor zones. Therefore, although the inter-reactor mass flow rates are extracted from CFD, the reactor networks remain strongly reduced by collapsing the fully resolved reacting flow into a small number of idealized reactor zones with fixed inter-zone exchange. Also, this two-level clustering approach addresses two limitations of a direct RL-based mapping. First, it avoids the memory burden associated with operating directly on all $10^6$ CFD points. Second, it promotes smoother and more physically coherent zone boundaries, whereas direct RL-based clustering produced irregular and noisy partitions. For comparison in Section~\ref{sec:result}, a baseline reactor network is separately constructed by applying $k$-means directly to the CFD dataset with $n_c = 7$, yielding 7 reactor zones in a single step without the intermediate micro-cluster stage. This purely $k$-means-derived network serves as the distance-based reference against which the RL-augmented network (Section~\ref{sec:clustering_rl}) is compared.

\subsubsection{Reinforcement-Learning Cluster Refinement}
\label{sec:clustering_rl}
Since $k$-means clustering is purely distance-based, it can group together regions with dissimilar thermochemical behavior, thereby reducing the physical consistency of the reactor network (RN) model. To address this limitation, the $k$-means micro-clusters were used as the initial partition, after which an \textit{actor--critic} RL approach \cite{konda1999actor} was employed to reassign the clusters for improved LBO prediction. At each iteration, given the current cluster-state descriptor $s_t$ (i.e., centroid thermochemical features), the actor network $\pi_{\theta}(a_t \mid s_t)$, parameterized by $\theta$, outputs a stochastic policy over possible cluster reassignments $a_t$. In other words, the actor selects an action, corresponding to a reassignment of clusters, based on the current state. The actor is implemented as a fully connected neural network with two hidden layers of size 128 and ReLU activations. Its output layer contains seven neurons, corresponding to the possible cluster assignments. These outputs are converted through a softmax function into a categorical probability distribution over actions, from which an action is sampled during training to encourage exploration. During inference, the action with the highest probability is selected, producing a deterministic policy.

The critic, by contrast, evaluates the quality of the current state by estimating the expected cumulative future reward associated with it. Specifically, the critic network $V_{\psi}(s_t)$, parameterized by $\psi$, approximates the state-value function and provides a baseline for assessing whether the actor's chosen action leads to improvement. The critic uses the same hidden-layer architecture as the actor. Together, the actor and critic are trained to refine the cluster boundaries in a manner that improves the predictive accuracy of the RN model for lean blowout. At inference, only the trained actor is retained. During training, we characterize lean blowout based on the maximum reactor temperature falling below a critical temperature, $T_c$. $T_c = 1300$ K is obtained by starting from a stable condition and progressively reducing $\phi$ while monitoring the maximum reactor temperature in the CRN. Near extinction, a rapid collapse in $T_{\max}$ is observed, indicating that stable combustion can no longer be sustained below this temperature threshold (as shown in Fig.~S2). Using this temperature-based criterion avoids performing full transient LBO calculations during RL training, which would be computationally prohibitive given the thousands of candidate reactor-network configurations explored.

\begin{figure}[H]
\centering
\includegraphics[width=0.5\textwidth]{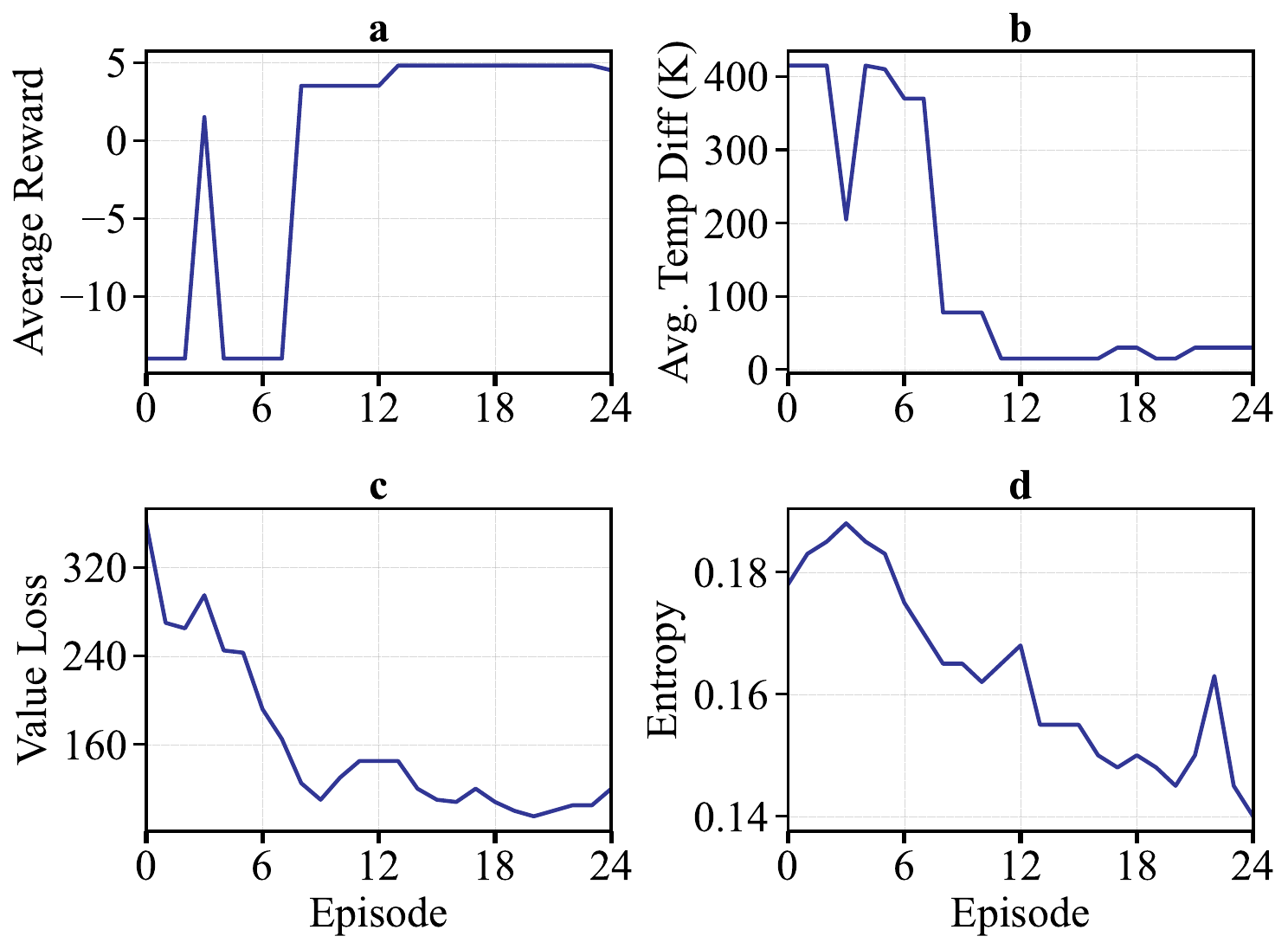}
\caption{RL training metrics as functions of training episode. One episode consists of 100 iterations.}
\label{fig:rl_results}
\end{figure}

During training, the RL procedure operates in reverse by fixing the reactor-network equivalence ratio at the LES-derived $\phi_{\mathrm{LBO}}$, which is obtained by running a single high fidelity LES. The objective is then to adjust the reactor partition such that, at this known blowout condition, the reactor-network prediction produces a maximum temperature as close as possible to the surrogate extinction marker, $T_c$. Therefore, after each reactor-network simulation step, a temperature-based reward that penalizes deviations from $T_c$ is computed as:
\begin{equation}
R_t = \exp\!\left(-\frac{|T_t - T_{\text{c}}|}{100}\right) - 1,
\label{eq:reward}
\end{equation}
The actor parameters are updated following the policy gradient:
\begin{equation}
\nabla_{\theta} J(\theta) = \mathbb{E}\big[\nabla_{\theta} \log \pi_{\theta}(a_t|s_t)\, \delta_t \big],
\label{eq:policy_gradient}
\end{equation}
where $\delta_t$ is the temporal difference error defined as:
\begin{equation}
\delta_t = R_t + \gamma V_{\psi}(s_{t+1}) - V_{\psi}(s_t),
\label{eq:td_error}
\end{equation}
while the critic minimizes the temporal-difference loss:
\begin{equation}
\mathcal{L}_V = \big(R_t + \gamma V_{\psi}(s_{t+1}) - V_{\psi}(s_t)\big)^2.
\label{eq:value_loss}
\end{equation}
An additional entropy regularization term $\mathcal{L}_H = -\sum_a \pi_{\theta}(a_t|s_t)\log\pi_{\theta}(a_t|s_t)$ is applied to encourage exploration. The overall optimization objective combines the policy-gradient objective, the value-function loss, and an entropy regularization term:
\begin{equation}
\mathcal{L}_{\text{total}} = -J(\theta) + \lambda_V \mathcal{L}_V - \lambda_H \mathcal{L}_H,
\label{eq:total_loss}
\end{equation}
where $\lambda_V$ and $\lambda_H$ are weighting coefficients for stability. Training employed discount factor $\gamma = 0.99$, $\lambda_V = 0.5$, and $\lambda_H$ initially set to $0.01$ and annealed exponentially to encourage policy convergence. These values are in line with commonly used hyperparameter settings reported in the literature \cite{schulman2017ppo,gleave2021epic,petrazzini2021beta}. From Fig.~\ref{fig:rl_results}, we see that during training, the RL agent progressively reduces the mean temperature deviation across episodes, as seen by increasing reward and decreasing entropy over the training horizon. After training, the reactor network was deployed to predict $\phi_{\textrm{LBO}}$ using a staged equivalence-ratio reduction procedure analogous to the CFD approach described later in Section~\ref{sec:transient_lbo}. Starting from a stable operating condition, successive steady-state calculations were performed at progressively lower equivalence ratios until $T_{\max}<T_c$, which defines blowout during evaluation. During training, however, the target $\phi_{\textrm{LBO}}$ is already known, so Eq.~\ref{eq:reward} penalizes deviations from $T_c$ on either side, since temperatures above or below $T_c$ correspond to delayed or premature blowout prediction, respectively.

\section{Results}
\label{sec:result}

\subsection{CFD Flame Structure}
\label{sec:cfd_results}
\begin{figure}[H]
\centering
\includegraphics[width=0.42\textwidth]{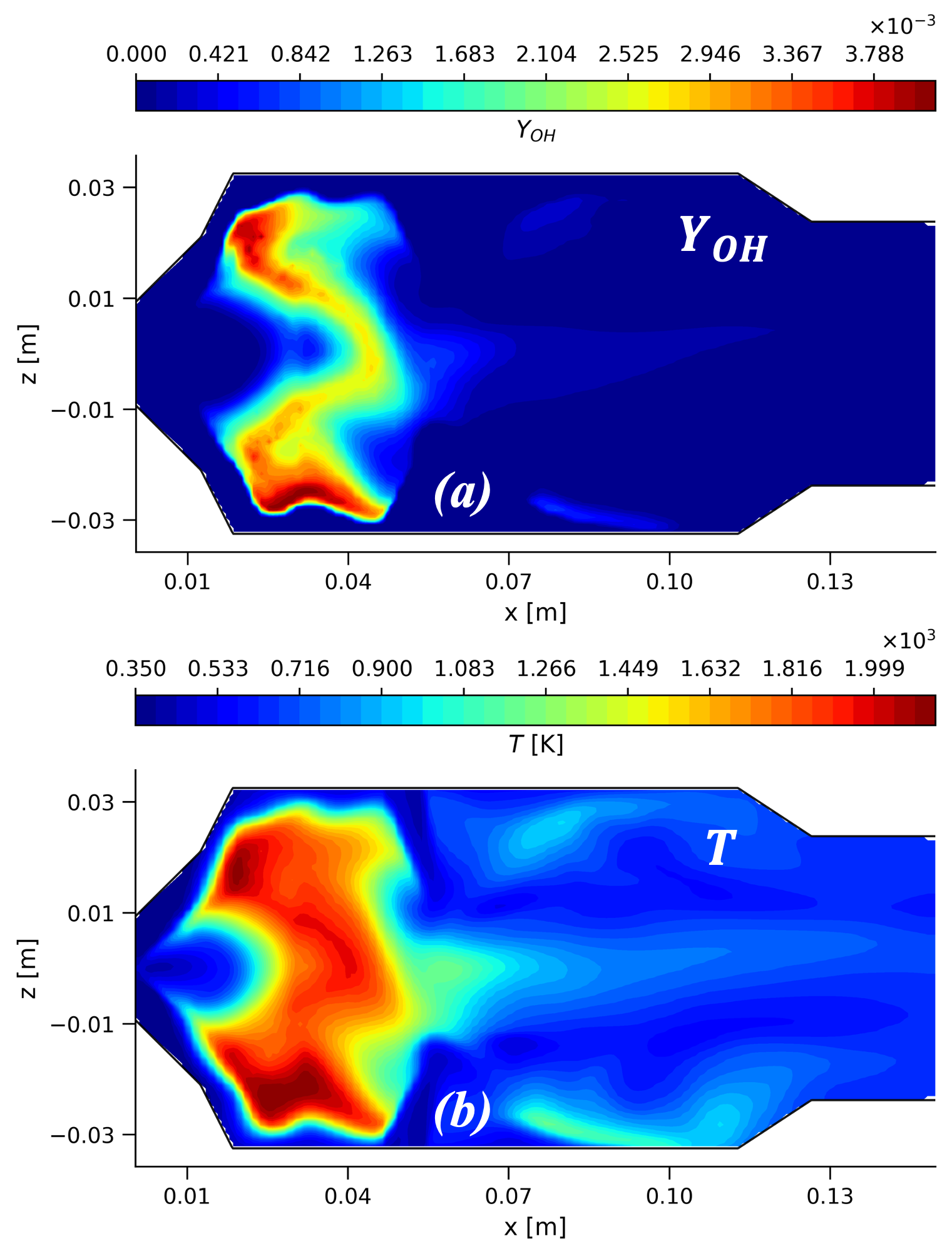}
\caption{(a) \ce{OH} mass fraction and (b) temperature contours obtained from the CFD solution at $\phi = 0.35$.}
\label{fig:contour}
\end{figure}

Fig.~\ref{fig:contour} presents the spatial distribution of \ce{OH} mass fraction and temperature on the $y = 0$ plane from the steady-state CFD solution at the reference operating condition ($\phi = 0.35$). Peak \ce{OH} mass fractions of $3.8 \times 10^{-3}$ are concentrated within the central recirculation zone ($x = 0.02$--$0.04$~m), where droplet evaporation, mixing, and primary heat release occur simultaneously. Downstream of $x \approx 0.06$~m, \ce{OH} concentrations decay sharply, marking the completion of the principal heat release zone. The region of elevated temperature occupies a broader spatial volume, with peak values near 2000~K coinciding with the high-\ce{OH} region, yet retaining pockets of elevated temperature further downstream ($x > 0.08$~m) due to the continued oxidation of intermediate combustion products beyond the primary reaction zone. The spatial distributions of \ce{OH} mass fraction and temperature revealed by these contours serve as the primary thermochemical basis for reactor zone construction, and their structure directly influences the resulting cluster boundaries, as discussed in the following section.

\subsection{Reactor Zone Partitioning}
\label{sec:partitioning}
\begin{figure}[H]
\centering
\includegraphics[width=0.42\textwidth]{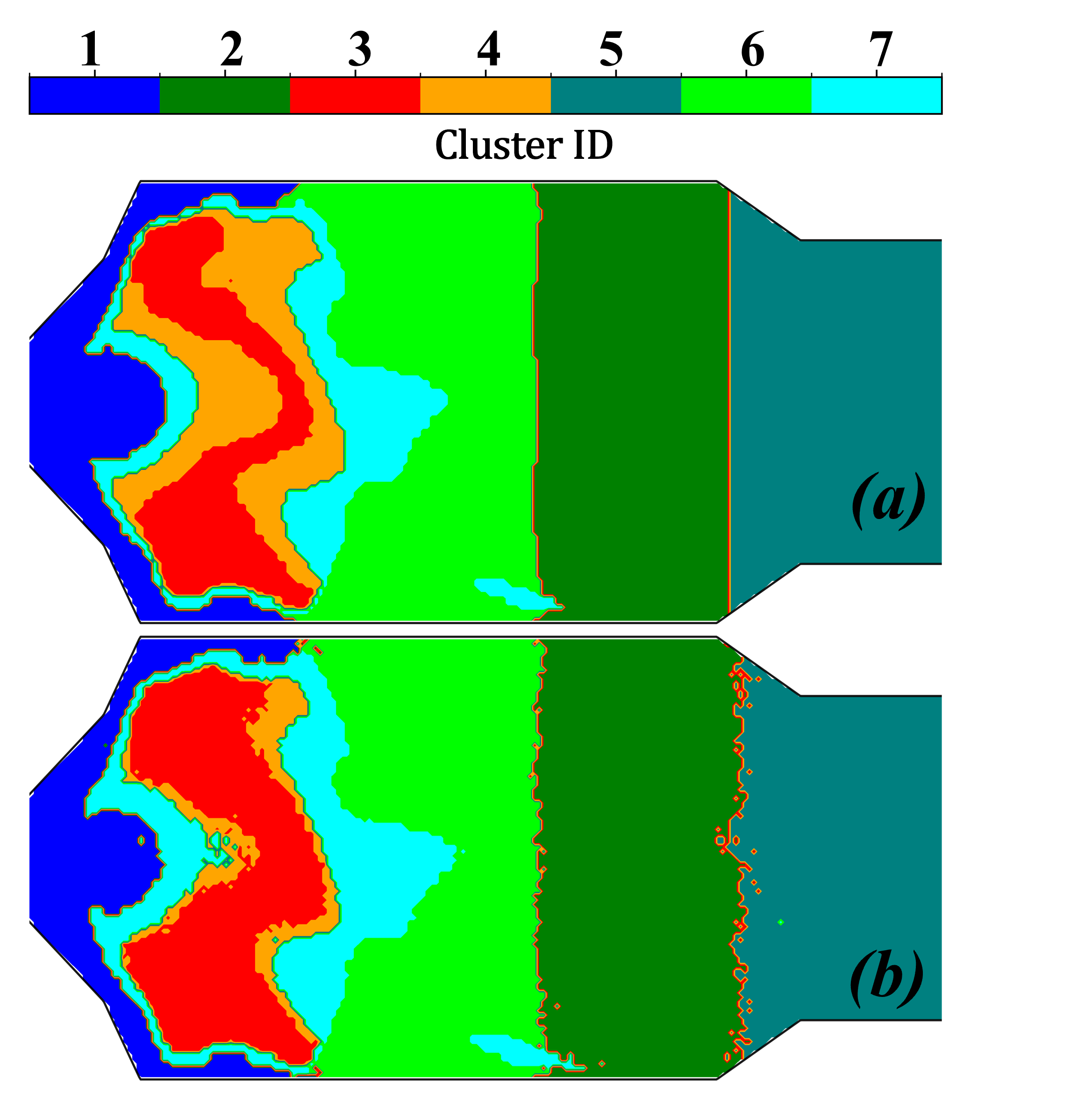}
\caption{Reactor zones obtained from (a) $k$-means clustering and (b) the RL-based partitioning.}
\label{fig:clusters}
\end{figure}

The seven clusters in Fig.~\ref{fig:clusters} correspond to the following reactor types: an evaporator/breakup PSR immediately downstream of the injector (cluster/reactor 1); a mixing PSR after this region to capture continued evaporation (cluster/reactor 7); four standard PSRs representing the 2 flame zones (reactors 3 and 4), recirculation (reactor 6), and dilution regions (reactor 2); and a downstream post-flame PFR (reactor 5). The reactor interconnections are shown schematically in Fig.~S1 of the Supplemental Material. The reactor assignments follow the underlying flow physics: zones characterized by strong transport effects are modeled as PSRs, while the downstream predominantly axial post-flame region is modeled as a PFR. The evaporator/breakup and mixing reactors with modifications to capture breakup or evaporation are placed downstream of the injector to capture primary atomization, evaporation, and fuel-air mixing prior to combustion.

Fig.~\ref{fig:clusters} compares the domain partitioning produced by the $k$-means algorithm and the RL policy for a seven-reactor network. The most consequential difference between the two partitionings concerns Reactor Zone~3, which largely overlaps with the primary combustion region identified in Fig.~\ref{fig:contour}. The $k$-means algorithm, operating purely on Euclidean distance in the feature space, assigns Zone~3 a volume of approximately 28~cm$^3$. This compact zone captures only the region of highest \ce{OH} concentration and temperature, which is essentially the core of the primary flame region. The RL policy, in contrast, expands Zone~3 to approximately 39~cm$^3$, a 39\% increase in volume relative to the $k$-means partition. This broader zone encompasses not only the peak-reactivity core but also the surrounding region of moderate heat release, including portions of the recirculation zone where \ce{OH} is elevated but not at its maximum.

These differences between Figs.~\ref{fig:clusters}a and b arise from a fundamental difference between the two clustering objectives. The $k$-means algorithm minimizes intra-cluster variance across all spatial features uniformly, which naturally biases it toward clusters with homogeneous representations. In contrast, the RL policy is driven exclusively by a reward signal tied to the accuracy of LBO prediction. Since LBO is governed by the global extinction behavior of the primary reaction zone, the RL agent learns to include the moderate-reactivity periphery of the flame within Zone~3. This has significant implications for the predictive accuracy of both approaches, as will be discussed in subsequent sections. Downstream zones that are less chemically active (Reactors 2, 5, and 6) remain largely unaffected by the RL optimization, consistent with their limited influence on the LBO reward signal.

\subsection{Transient LBO Simulation}
\label{sec:transient_lbo}
\begin{figure}[H]
\centering
\includegraphics[width=0.5\textwidth]{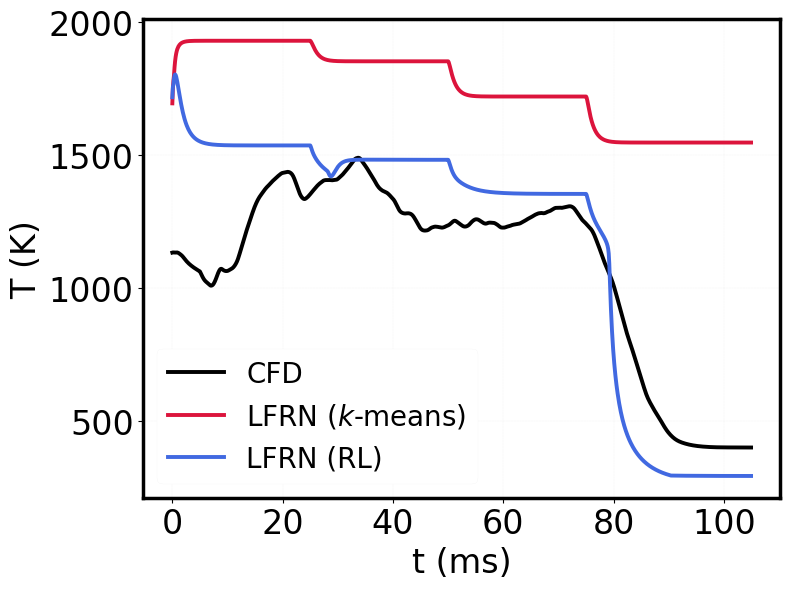}
\caption{Volume-averaged primary-zone temperature during transient ramp-down toward LBO.}
\label{fig:transient_temp}
\end{figure}

A transient lean blowout simulation was conducted to validate the framework under dynamic operating conditions. The vicinity of the LBO equivalence ratio was first identified using a coarse staged reduction of the fuel flow rate, starting from $\phi \approx 0.35$ and reducing the equivalence ratio in decrements of $\Delta \phi = 0.0435$ until extinction occurred. For the baseline case, three coarse staged reductions were required to reach the vicinity of the LBO condition. The equivalence ratio was then further refined using smaller decrements of $\Delta \phi = 0.001$ to determine the final reported $\phi_{\text{LBO}}$ more accurately. The reported LBO values correspond to the fine-step refinement rather than the initial coarse staging procedure. The following results in this subsection and thereafter should be interpreted in the context that the RL-derived reactor network is informed by data from a baseline LES, while the $k$-means only includes information from RANS. The objective is therefore to assess the added improvements obtained from a goal-oriented partitioning strategy, but this involves an additional upfront cost of a baseline LES.

Fig.~\ref{fig:transient_temp} presents the volume-averaged mean temperature as a function of time for the CFD reference, the RL-based LFRN and the $k$-means LFRN. All three results are averaged over the same fixed control volume ($x \in [0,\,0.05]$\,m, $y,z \in [-0.025,\,0.025]$\,m) enclosing the primary reaction zone. The CFD result is smoothed with a 10\,ms moving average to suppress LES oscillations, and the early-time dip in temperature is a numerical artifact due to the presence of a coarse mesh in the domain before the full activation of AMR. The CFD shows a gradual thermal decline with decreasing $\phi$, followed by a sharp collapse at $\tau_{\text{LBO}} \approx 80$\,ms ($\phi \approx 0.219$). The RL-based LFRN reproduces this behavior closely, predicting blowout at nearly the same $\phi$, although with a slightly earlier temperature collapse due to the reduced thermal inertia of the seven-reactor representation. The $k$-means-based LFRN, by contrast, fails to predict blowout entirely, with the volume-averaged temperature remaining elevated at approximately 1600~K throughout the full equivalence ratio ramp, including at the final condition where both the CFD and the RL-based network indicate global extinction.

\subsection{Timescale Analysis of LBO}
\label{sec:timescale}
To explain the divergent LBO predictions of the RL-based and $k$-means reactor networks, a timescale analysis was performed for Zone~3, identified in Section~\ref{sec:cfd_results} as the primary combustion zone and region of highest \ce{OH} activity. Three representative times were considered: $\tau \approx 0.35 \tau_{\text{LBO}}$ (far from LBO), $\tau \approx 0.8 \tau_{\text{LBO}}$ (near LBO), and $\tau \approx \tau_{\text{LBO}}$ (at LBO). The flow timescale was defined as $\tau_{\text{flow}} = \bar{\rho}V/\dot{m}$, where $\bar{\rho}$ is the zone-averaged density, $V$ is the zone volume, and $\dot{m}$ is the mass flow rate through the zone. The chemical timescale was taken as the residence time at blowout, $\tau_{\text{chem}} = \tau_{\text{res}}$, similar to Wang et al.~\cite{WANG2021106532}, and the Damk\"{o}hler number was defined as $Da = \tau_{\text{flow}}/\tau_{\text{chem}}$.

\begin{figure}[H]
\centering
\includegraphics[width=0.5\textwidth]{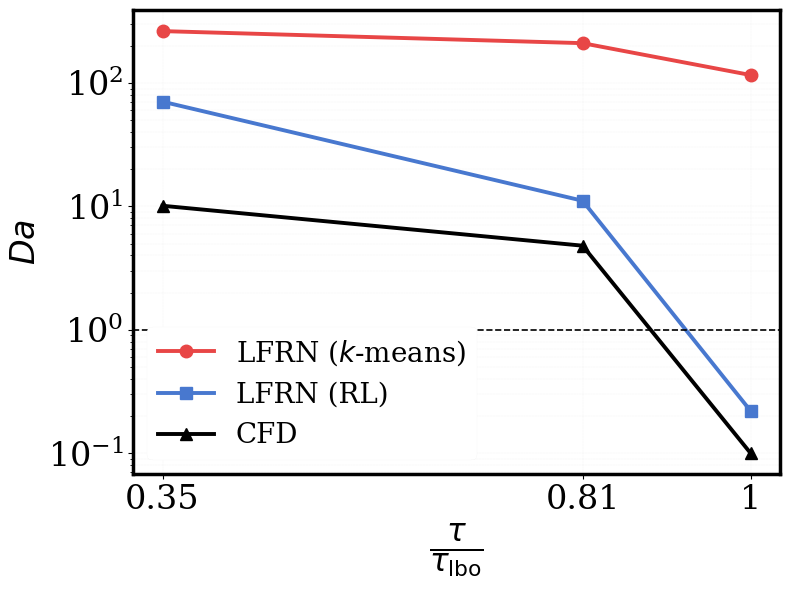}
\caption{Damk\"{o}hler number in Zone~3 at three representative timesteps for the CFD reference, LFRN (RL), and LFRN ($k$-means). The dashed line indicates $Da = 1$.}
\label{fig:damkholer}
\end{figure}

We emphasize that the purpose of this analysis is not to directly compare the individual timescales of the LES and reactor-network models. The LES resolves fully unsteady three-dimensional transport, whereas the reactor network represents coarse averaged inter-zone exchange, making direct comparison of timescales (especially flow) across frameworks of limited physical consistency. This mismatch is inherent to coarse CRN models constructed from RANS mean fields, which cannot resolve the instantaneous turbulent fluctuations and recirculation dynamics associated with transient LBO. Consequently, the reactor network evolves through a sequence of quasi-steady states as the equivalence ratio is reduced. Therefore, the analysis focuses on whether the ratio of flow and chemical timescales within each framework remains internally consistent as blowout is approached, and on why the RL-derived partition succeeds while the $k$-means partition does not.

Figure~\ref{fig:damkholer} summarizes the resulting Damk\"{o}hler-number evolution, with details about the individual timescales shown in Table S1 in the Supplemental Material. Far from LBO ($\tau \approx 0.35\tau_{\text{LBO}}$), all models remain in a stable combustion regime with $Da \gg 1$. However, the RL and $k$-means networks already differ substantially in their chemical timescales. The RL-derived Zone~3 predicts $\tau_{\text{chem}} \approx 4.72~\mu$s. This difference arises because the $k$-means partition isolates only the peak-reactivity core of the flame, producing artificially high zone-averaged temperature and \ce{OH} concentration, and therefore unrealistically short chemical timescales. As the equivalence ratio is reduced toward blowout, the divergence between the two reactor networks becomes more pronounced. Near LBO ($\tau \approx 0.8\tau_{\text{LBO}}$), the RL-derived network predicts a substantial increase in chemical timescale to approximately $27.9~\mu$s. In contrast, the $k$-means network still predicts a short chemical timescale of only $1.15~\mu$s, indicating that its Zone~3 remains strongly reactive despite the weakening global flame. Consequently, the RL-derived Damk\"{o}hler number decreases significantly as extinction is approached, while the $k$-means-derived value remains artificially large.

Finally, at $\tau \approx \tau_{\text{LBO}}$, both the CFD and RL-derived reactor network exhibit a sharp increase in chemical timescale, reaching approximately $1020~\mu$s and $1300~\mu$s, respectively. In both cases, this produces $Da < 1$ ($Da_{\text{CFD}} \approx 0.10$, $Da_{\text{RL}} \approx 0.22$), indicating that the reacting mixture is no longer able to sustain combustion before being convected from the primary zone. The $k$-means network, however, continues to predict a very short chemical timescale ($\tau_{\text{chem}} \approx 2.22~\mu$s) and a large Damk\"{o}hler number ($Da \approx 116$), showing no indication of impending blowout. These results suggest that the RL model succeeds because, by learning from experience informed by the reward signal in Eq.~\ref{eq:reward}, it learns reactor boundaries that are consistent with the observed blowout. In particular, the RL-derived partition captures the sharp increase in $\tau_{\text{chem}}$ near extinction, causing the ratio $\tau_{\text{flow}}/\tau_{\text{chem}}$ to evolve in a manner qualitatively consistent with the CFD behavior, as shown in Fig.~\ref{fig:damkholer}.

\subsection{Sensitivity to Number of Micro-Clusters}
\label{sec:nf_sensitivity}
\begin{figure}[H]
\centering
\includegraphics[width=0.45\textwidth]{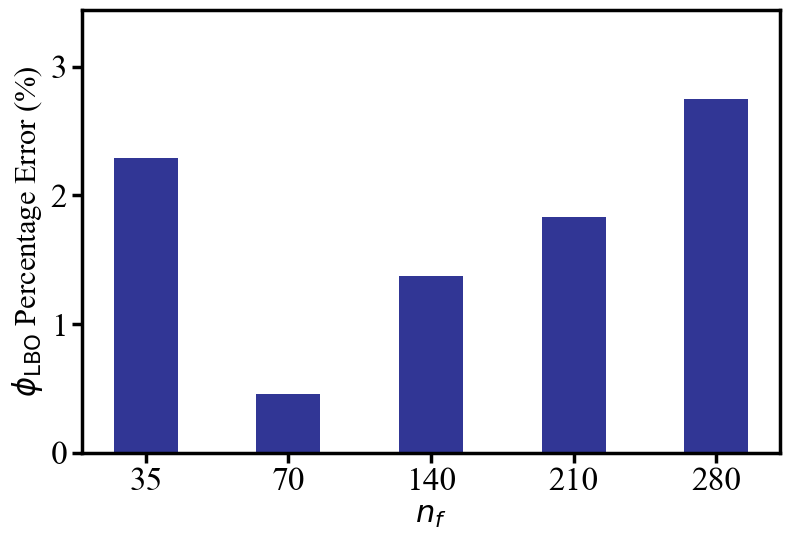}
\caption{Percentage error in $\phi_{\text{LBO}}$ prediction as a function of the number of micro-clusters $n_f$.}
\label{fig:nf_error}
\end{figure}

A sensitivity study was conducted by varying the number of micro-clusters $n_f$ over the range 35--280 while keeping the number of coarse reactor zones fixed at $n_c = 7$. Figure~\ref{fig:nf_error} summarizes the resulting error in $\phi_{\text{LBO}}$ relative to the CFD reference value ($\phi_{\text{LBO}} = 0.218$). The results show that $n_f = 70$ yields the lowest prediction error of 0.46\%. Very small values of $n_f$ lead to higher errors due to coarser zone boundaries, since the number of micro-cluster centroids would not provide the RL algorithm with sufficient degrees of freedom to precisely capture the optimal reactor boundaries. Conversely, at larger values of $n_f$, the performance deteriorates as the increased number of micro-clusters introduces noisier and less coherent zone boundaries. Despite this deterioration at both extremes, the errors remain relatively small across the range of $n_f$ values considered, which spans a broad eightfold increase from $n_f = 35$ to $n_f = 280$.

\subsection{LBO Sensitivity to Primary Air Flow Rate and Inlet Temperature}
\label{sec:lbo_Air}
\begin{figure}[H]
\centering
\includegraphics[width=0.5\textwidth]{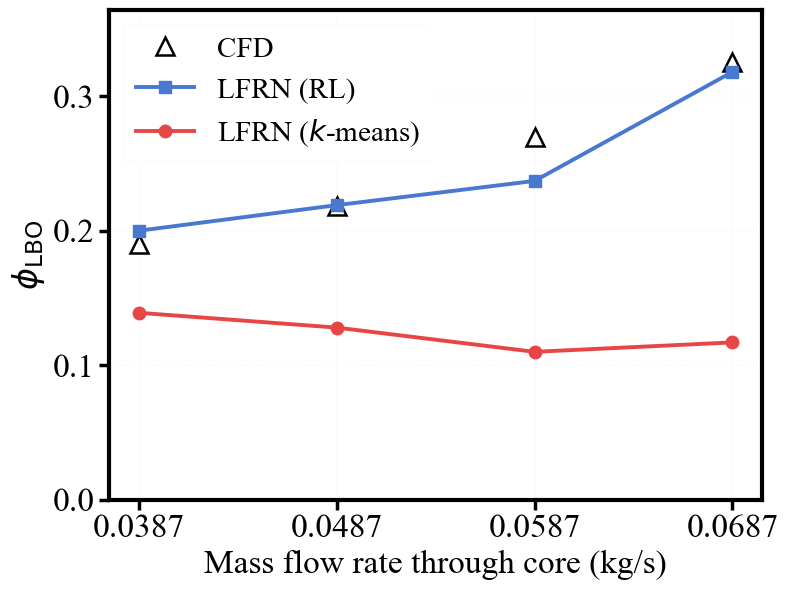}
\caption{Parametric effect of mass flow rate on $\phi_{\text{LBO}}$.}
\label{fig:lbo_comparison}
\end{figure}

A parametric study was conducted to examine the effect of the primary air mass flow rate on LBO limits. Fig.~\ref{fig:lbo_comparison} compares LBO predictions from both reactor-network approaches against CFD simulations across a primary mass flow rate range of 0.0387--0.0687~kg/s, which corresponds to primary equivalence ratios of 0.219 to 0.325 at the baseline operating condition. The result from the 0.0487~kg/s air flow rate (baseline) was used to train the RL-driven framework. The resulting network was then applied to the remaining operating conditions within the mass flow rate range. The maximum mass flow rate corresponds to an increase of approximately 77\% relative to the minimum value considered.

The CFD results demonstrate a positive correlation between the primary air mass flow rate and $\phi_{\text{LBO}}$, with values ranging from 0.219 to 0.325. This behavior is physically consistent. At fixed fuel flow, increasing $\dot{m}$ shortens the residence time in the primary zone, since $\tau_{\text{flow}} \propto \rho V/\dot{m}$, thereby also lowering the Damk\"ohler number, $\text{Da} = \tau_{\text{flow}}/\tau_{\text{chem}}$. As the mixture is leaned toward LBO, $\tau_{\text{chem}}$ rises sharply (Section~\ref{sec:timescale}), so $\text{Da}$ falls below unity at a richer mixture than it would at the baseline flow rate, shifting the LBO point to a higher $\phi$. Increased turbulent mixing at higher $\dot{m}$ reinforces this trend by raising local strain rates and entraining cooler unreacted gas into the reaction zone, but this is a secondary effect. The $k$-means approach, relying solely on distances within the input feature space, is unable to capture this trend, resulting in a non-monotonic variation of $\phi_{\text{LBO}}$ with increasing mass flow rate (i.e., a decrease, followed by a slight increase). In contrast, the RL model captures the correct qualitative trend of increasing $\phi_{\text{LBO}}$ with increasing mass flow rate across the entire range of conditions. At the baseline condition, unsurprisingly, the RL model produces a very low error of approximately 0.46\%, confirming that the RL agent successfully identified near-optimal cluster boundaries. At non-baseline conditions, higher quantitative errors are observed. The sensitivity of $\phi_{\text{LBO}}$ with respect to the inlet mass flow rate at the reference condition is lower compared to the CFD solution, as evidenced by the lower slope. This results in an overprediction of $\phi_{\text{LBO}}$ at 0.0487~kg/s and an underprediction at 0.0587~kg/s. However, the error levels remain acceptable given the low complexity of a seven-reactor reduced-order network.

\begin{figure}[H]
\centering
\includegraphics[width=0.5\textwidth]{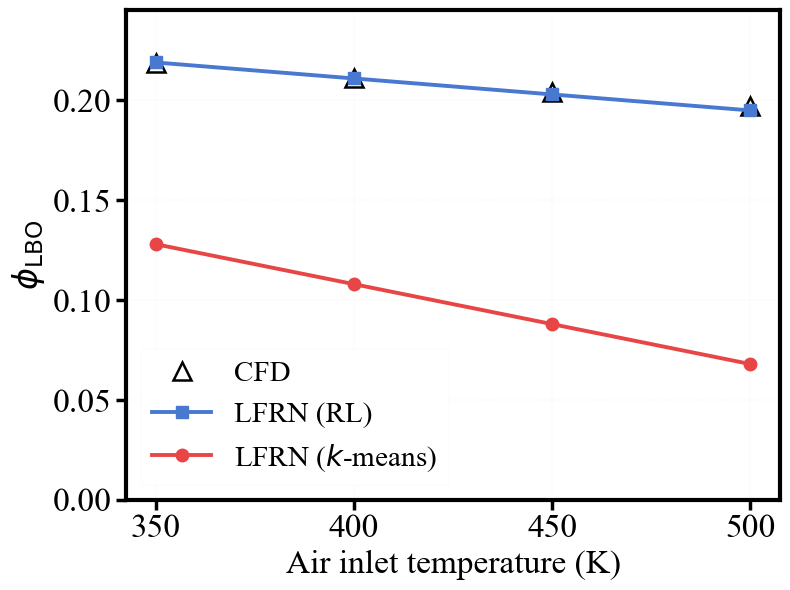}
\caption{Parametric effect of air inlet temperature on $\phi_{\text{LBO}}$.}
\label{fig:temp_comparison}
\end{figure}

A second parametric study investigated variation in $\phi_{\text{LBO}}$ as the inlet temperature is varied from 350~K to 500~K. Fig.~\ref{fig:temp_comparison} presents the LBO predictions from both reactor network approaches compared to the CFD reference. The CFD results show that increasing inlet temperature from 350~K to 500~K reduces $\phi_{\text{LBO}}$ from 0.2187 to 0.197. This trend is physically consistent, since higher inlet temperatures lead to increased reactivity of the mixture and shortens the chemical timescale in the primary zone. As a result, combustion can be sustained for longer as the mixture is leaned. The RL approach demonstrates a high level of accuracy, capturing both the magnitude and direction of this trend with errors below 0.14\% at both conditions. In contrast, the $k$-means approach exhibits substantial under-predictions of 41.47\% and 65.48\% at 350~K and 500~K, respectively, though it qualitatively captures the decreasing trend in $\phi_{\text{LBO}}$.

\subsection{Computational Efficiency}
\label{sec:cost}
The LES computations employed in this study required 256 processors and 2--4 days per operating condition. In contrast, the reactor network predicted $\phi_{\text{LBO}}$ in approximately 22~s on a 12-core Intel Core i7 workstation, corresponding to a speedup exceeding 7{,}800 while retaining the correct qualitative LBO trends. The entire RL training procedure completed in approximately 3~h. It should be noted that this speedup estimation does not include the initial LES cost required to generate the baseline training data. It only captures the reduced cost of evaluating additional operating conditions once the baseline CFD-informed model has been constructed.

\section{Conclusions}
\label{sec:conclusion}
This study presented a reinforcement-learning-augmented liquid-fueled reactor network framework for predicting lean blowout in gas turbine combustors using goal-oriented reactor partitioning. The results show that LBO prediction depends strongly on how the domain is partitioned, and that goal-oriented zoning provides a more effective alternative to conventional distance-based clustering. In practice, the lightweight liquid fueled reactor can be used to reduce the number of expensive CFD simulations required during combustor design by rapidly screening candidate operating conditions, identifying promising regions of the design space, and guiding targeted high-fidelity simulations. Future work will extend the framework to additional fuels and combustor configurations and incorporate \ce{NOx} prediction into the RL reward formulation.

\section*{CRediT authorship contribution statement}
\textbf{Philip John}: Data curation, Formal analysis, Investigation, Methodology, Software, Visualization, Writing -- original draft.
\textbf{Eloghosa Ikponmwoba}: Methodology, Software, Formal analysis, Investigation, Writing -- original draft.
\textbf{Pinaki Pal}: Methodology, Writing -- review \& editing.
\textbf{Opeoluwa Owoyele}: Conceptualization, Funding acquisition, Methodology, Supervision, Project administration, Resources, Writing -- review \& editing.

\section*{Declaration of competing interest}
The authors declare that they have no known competing financial interests or personal relationships that could have appeared to influence the work reported in this paper.

\section*{Acknowledgments}
This material is based upon work supported by the U.S. Department of Energy, Office of Energy Efficiency and Renewable Energy, Bioenergy Technologies Office under Award Number DE-SC0023463. The CONVERGE software licenses used in this work were provided through the CONVERGE Academic Program, and Convergent Science provided the baseline geometry utilized in this study. Portions of this research were conducted with high performance computational resources provided by Louisiana State University (\url{http://www.hpc.lsu.edu}). The submitted manuscript has been partly created by UChicago Argonne, LLC, Operator of Argonne National Laboratory (Argonne). Argonne, a U.S. Department of Energy Office of Science Laboratory, operated under Contract No. DE-AC02-06CH11357.

\FloatBarrier

\bibliographystyle{unsrt}
\bibliography{references}

\end{document}